\documentclass[11pt,letterpaper]{article}
\usepackage{emnlp2017}
\usepackage{times}
\usepackage{latexsym}

\usepackage{url}

\usepackage{graphicx}
\usepackage{amsmath}
\usepackage{amsfonts}
\usepackage{algorithm}
\usepackage{algorithmic}

\usepackage{float}
\usepackage{multirow}
\usepackage{verbatim}
\usepackage{caption}
\usepackage{subcaption}

\emnlpfinalcopy

\def\emnlppaperid{550}

\title{Deep Recurrent Generative Decoder for Abstractive Text Summarization\Thanks{The work described in this paper is supported by a grant from the Grant Council of the Hong Kong Special Administrative Region, China (Project Code: 14203414).}}
	
\author{ Piji Li$^{\dag}$ \ \ Wai Lam$^{\dag}$ \ \ Lidong Bing$^{\ddag}$ \ \ Zihao Wang$^{\dag}$ \\
		$^{\dag}$Key Laboratory on High Confidence Software Technologies (Sub-Lab, CUHK), \\Ministry of Education, China\\
		$^{\dag}$Department of Systems Engineering and Engineering Management,\\
		The Chinese University of Hong Kong\\
		$^{\ddag}$AI Lab, Tencent Inc., Shenzhen, China\\
		{\tt  $^{\dag}$\{pjli, wlam, zhwang\}@se.cuhk.edu.hk, $^{\ddag}$lyndonbing@tencent.com}}

\date{}

\begin{document}

\maketitle

\begin{abstract}
  We propose a new framework for abstractive text summarization based on a sequence-to-sequence oriented encoder-decoder model equipped with a deep recurrent generative decoder (DRGN).
  Latent structure information implied in the target summaries is learned based on a recurrent latent random model for improving the summarization quality.
  Neural variational inference is employed to address the intractable posterior inference for the recurrent latent variables.
  Abstractive summaries are generated based on both the generative latent variables and the discriminative deterministic states.
  Extensive experiments on some benchmark datasets in different languages show that DRGN achieves improvements over the state-of-the-art methods.
\end{abstract}

\section{Introduction}
\label{sec:1}

Automatic summarization is the process of automatically generating a summary that retains the most important content of the original text document \cite{edmundson1969new,luhn1958automatic,nenkova2012survey}.
Different from the common extraction-based and compression-based methods, abstraction-based methods aim at constructing new sentences as summaries, thus they require a deeper understanding of the text and the capability of generating new sentences, which provide an obvious advantage in improving the focus of a summary, reducing the redundancy, and keeping a good compression rate \cite{lidong15absmds,rush2015neural,nallapati2016abstractive}.

\begin{figure}[!t]
	\centering
	\includegraphics[width=0.9\columnwidth]{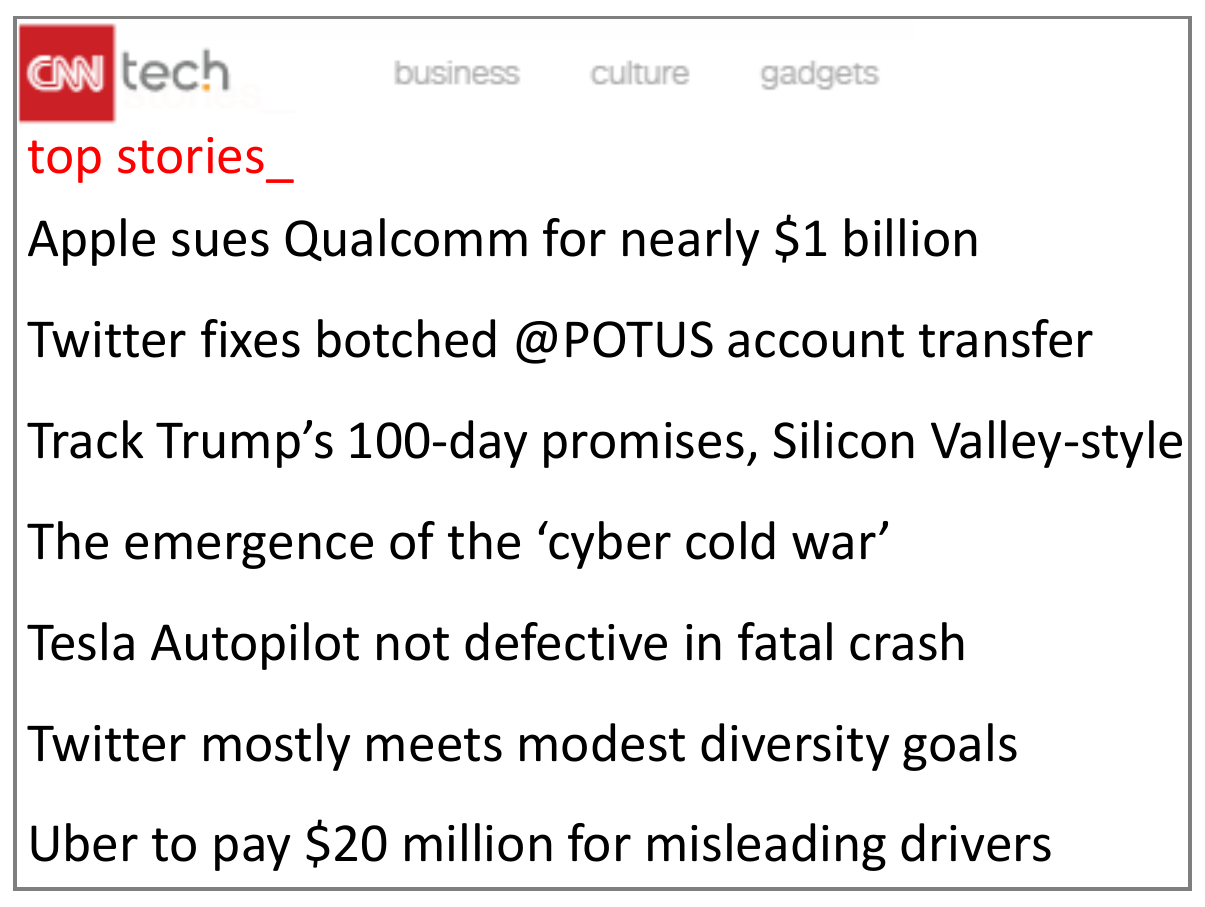}
	\caption{
		Headlines of the top stories from the channel ``Technology'' of CNN.
	}
	\label{fig:front}
\end{figure}

Some previous research works show that human-written summaries are more abstractive \cite{jing2000cut}.
Moreover, our investigation reveals that people may naturally follow some inherent structures when they write the abstractive summaries.
To illustrate this observation, we show some examples in Figure~\ref{fig:front}, which are some top story summaries or headlines from the channel ``Technology'' of CNN. 
After analyzing the summaries carefully, we can find some common structures from them, such as ``\textbf{What}'', ``\textbf{What-Happened}'' , ``\textbf{Who Action What}'', etc.
For example, the summary ``\textit{Apple sues Qualcomm for nearly \$1 billion}'' can be  structuralized as ``Who (Apple) Action (sues) What (Qualcomm)''.
Similarly, the summaries ``\textit{[Twitter] [fixes] [botched @POTUS account transfer]}'',  ``\textit{[Uber] [to pay] [\$20 million] for misleading drivers}'', and ``\textit{[Bipartisan bill] aims to [reform] [H-1B visa system]}'' also follow the structure of ``Who Action What''.
The summary ``\textit{The emergence of the `cyber cold war}''' matches with the structure of ``What'', and the summary ``\textit{St. Louis' public library computers hacked}'' follows the structure of ``What-Happened''.

Intuitively, if we can incorporate the latent structure information of summaries into the abstractive summarization model, it will improve the quality of the generated summaries.
However, very few existing works specifically consider the latent structure information of summaries in their summarization models.
Although a very popular neural network based sequence-to-sequence (seq2seq) framework has been proposed to tackle the abstractive summarization problem \cite{lopyrev2015generating,rush2015neural,nallapati2016abstractive}, the calculation of the internal decoding states is entirely deterministic.
The deterministic transformations in these discriminative models lead to limitations on the representation ability of the latent structure information.
\citet{miao2016language} extended the seq2seq framework and proposed a generative model to capture the latent summary information, but they did not consider the recurrent dependencies in their generative model leading to limited representation ability.

To tackle the above mentioned problems, we design a new framework based on sequence-to-sequence oriented encoder-decoder model equipped with a latent structure modeling component.  
We employ Variational Auto-Encoders (VAEs) \cite{kingma2013auto,rezende2014stochastic} as the base model for our generative framework which can handle the inference problem associated with complex generative modeling.
However, the standard framework of VAEs is not designed for sequence modeling related tasks.
Inspired by \cite{chung2015recurrent}, we add historical dependencies on the latent variables of VAEs and propose a deep recurrent generative decoder (DRGD) for latent structure modeling.
Then the standard discriminative deterministic decoder and the recurrent generative decoder are integrated into a unified decoding framework.
The target summaries will be decoded based on both the discriminative deterministic variables and the generative latent structural information.
All the neural parameters are learned by back-propagation in an end-to-end training paradigm.

The main contributions of our framework are summarized as follows:
(1) We propose a sequence-to-sequence oriented encoder-decoder model equipped with a deep recurrent generative decoder (DRGD) to model and learn the latent structure information implied in the target summaries of the training data. Neural variational inference is employed to address the intractable posterior inference for the recurrent latent variables.
(2) Both the generative latent structural information and the discriminative deterministic variables are jointly considered in the generation process of the abstractive summaries.
(3) Experimental results on some benchmark datasets in different languages show that our framework achieves better performance than the state-of-the-art models.

\section{Related Works}
Automatic summarization is the process of automatically generating a summary that retains the most important content of the original text document \cite{nenkova2012survey}.
Traditionally, the summarization methods can be classified into three categories:  extraction-based methods \cite{erkan2004lexrank,goldstein2000multi,wan2007manifold,min2012exploiting,nallapati2016summarunner,cheng2016neural,cao2016attsum,song2017summarizing}, compression-based methods \cite{li2013document,wang2013sentence,li2015reader,li2017salience}, and abstraction-based methods.
In fact, previous investigations show that human-written summaries are more abstractive \cite{barzilay2005sentence,lidong15absmds}.
Abstraction-based approaches can generate new sentences based on the facts from different source sentences.
\citet{barzilay2005sentence} employed sentence fusion to generate
a new sentence. \citet{lidong15absmds} proposed a more fine-grained fusion framework, where new sentences are generated by selecting and merging salient phrases.
These methods can be regarded as a kind of indirect abstractive summarization, and complicated constraints are used to guarantee the linguistic quality.

Recently, some researchers employ neural network based framework to tackle the abstractive summarization problem.
\citet{rush2015neural} proposed a neural network based model with local attention modeling, which is trained on the Gigaword corpus, but combined with an additional log-linear extractive summarization model with handcrafted features.
\citet{gu2016incorporating} integrated a copying mechanism into a seq2seq framework to improve the quality of the generated summaries.
\citet{chen2016distraction} proposed a new attention mechanism that not only considers the important source segments, but also distracts them in the decoding step in order to better grasp the overall meaning of input documents.
\citet{nallapati2016abstractive} utilized a trick to control the vocabulary size to improve the training efficiency. 
The calculations in these methods are all deterministic and the representation ability is limited.
\citet{miao2016language} extended the seq2seq framework and proposed a generative model to capture the latent summary information, but they do not consider the recurrent dependencies in their generative model leading to limited representation ability.

Some research works employ topic models to capture the latent information from source documents or sentences.
\citet{wang2009multi} proposed a new Bayesian sentence-based topic model by making use of both the term-document and term-sentence associations to improve the performance of sentence selection.
\citet{celikyilmaz2010hybrid} estimated scores for sentences based
on their latent characteristics using a hierarchical topic model, and trained a regression model to extract sentences.
However, they only use the latent topic information to conduct the sentence salience estimation for extractive summarization.
In contrast, our purpose is to model and learn the latent structure information from the target summaries and use it to enhance the performance of abstractive summarization.

\section{Framework Description}

\subsection{Overview}

\begin{figure*}[!t]
	\centering
	\includegraphics[width=2.0\columnwidth]{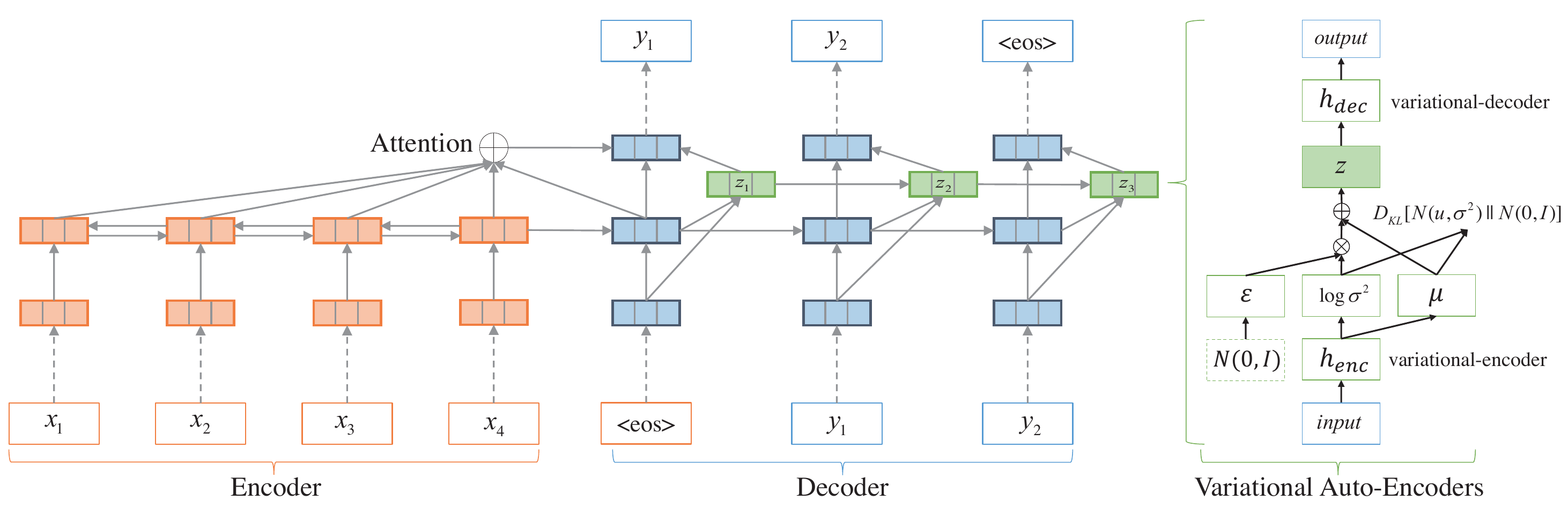}
	\caption{Our deep recurrent generative decoder (DRGD) for latent structure modeling.}
	\label{fig:framework}
\end{figure*}

As shown in Figure~\ref{fig:framework}, the basic framework of our approach is a neural network based encoder-decoder framework for sequence-to-sequence learning.
The input is a variable-length sequence ${X} = \{\mathbf{x}_1, \mathbf{x}_2, \ldots, \mathbf{x}_m\}$ representing the source text.
The word embedding $\mathbf{x}_t$ is initialized randomly and learned during the optimization process. 
The output is also a sequence ${Y} = \{\mathbf{y}_1, \mathbf{y}_2, \ldots, \mathbf{y}_n\}$, which represents the generated abstractive summaries.
Gated Recurrent Unit (GRU) \cite{cho2014learning} is employed as the basic sequence modeling component for the encoder and the decoder.
For latent structure modeling, 
we add historical dependencies on the latent variables of Variational Auto-Encoders (VAEs) and propose a deep recurrent generative decoder (DRGD) to distill the complex latent structures implied in the target summaries of the training data. 
Finally, the abstractive summaries will be decoded out based on both the discriminative deterministic variables ${H}$ and the generative latent structural information ${Z}$.

\subsection{Recurrent Generative Decoder}

Assume that we have obtained the source text representation $\mathbf{h}^e \in \mathbb{R}^{k_h}$.
The purpose of the decoder is to translate this source code $\mathbf{h}^e$ into a series of hidden states $\{\mathbf{h}^d_1, \mathbf{h}^d_2, \ldots, \mathbf{h}^d_n \}$, and then revert these hidden states to an actual word sequence and generate the summary.

For standard recurrent decoders, at each time step $t$, the hidden state $\mathbf{h}^d_t \in \mathbb{R}^{k_h}$ is calculated using the dependent input symbol $\mathbf{y}_{t-1} \in \mathbb{R}^{k_w}$ and the previous hidden state $\mathbf{h}^d_{t-1}$:
\begin{equation}
\mathbf{h}_t^d = f(\mathbf{y}_{t-1}, \mathbf{h}_{t-1}^d)
\end{equation}
where $f(\cdot)$ is a recurrent neural network such as vanilla RNN, Long Short-Term Memory (LSTM) \cite{hochreiter1997long}, and Gated Recurrent Unit (GRU) \cite{cho2014learning}. 
No matter which one we use for $f(\cdot)$, the common transformation operation is as follows:
\begin{equation}
\mathbf{h}_t^d = g(\mathbf{W}_{yh}^d\mathbf{y}_{t-1} + \mathbf{W}_{hh}^d\mathbf{h}_{t - 1}^d + \mathbf{b}_h^d)
\label{eq:deter_trans}
\end{equation}
where $\mathbf{W}_{yh}^d \in \mathbb{R} ^ {k_h \times k_w}$ and $\mathbf{W}_{hh}^d \in \mathbb{R} ^ {k_h \times k_h}$ are the linear transformation matrices. $\mathbf{b}_h^d$ is the bias. $k_h$ is the dimension of the hidden layers, and $k_w$ is the dimension of the word embeddings.
$g(\cdot)$ is the non-linear activation function.
From Equation~\ref{eq:deter_trans}, we can see that all the transformations are deterministic, which leads to a deterministic recurrent hidden state $h_t^d$.
From our investigations, we find that the representational power of such deterministic variables are limited.
Some more complex latent structures in the target summaries, such as the high-level syntactic features and latent topics, cannot be modeled effectively by the deterministic operations and variables.

Recently, a generative model called Variational Auto-Encoders (VAEs) \cite{kingma2013auto,rezende2014stochastic} shows strong capability in modeling latent random variables and improves the performance of tasks in different fields such as sentence generation \cite{bowman2016generating} and image generation \cite{gregor2015draw}.
However, the standard VAEs is not designed for modeling sequence directly.
Inspired by \cite{chung2015recurrent}, we extend the standard VAEs by introducing the historical latent variable dependencies to make it be capable of modeling sequence data.  
Our proposed latent structure modeling framework can be viewed as a sequence generative model which can be divided into two parts: inference (variational-encoder) and generation (variational-decoder). 
As shown in the decoder component of Figure~\ref{fig:framework}, the input of the original VAEs only contains the observed variable $\mathbf{y}_t$, and the variational-encoder can map it to a latent variable  $\mathbf{z} \in {\mathbb{R}^{k_z}}$, which can be used to reconstruct the original input.
For the task of summarization, in the sequence decoder component, the previous latent structure information needs to be considered for constructing more effective representations for the generation of the next state.

For the inference stage, the variational-encoder can map the observed variable $\mathbf{y}_{<t}$ and the previous latent structure information $\mathbf{z}_{<t}$  to the posterior probability distribution of the latent structure variable $p_\theta(\mathbf{z}_t|\mathbf{y}_{<t},\mathbf{z}_{<t})$.
It is obvious that this is a recurrent inference process in which $\mathbf{z}_t$ contains the historical dynamic latent structure information.
Compared with the variational inference process $p_\theta(\mathbf{z}_t|\mathbf{y}_{t})$ of the typical VAEs model,
the recurrent framework can extract more complex and effective latent structure features implied in the sequence data.   

For the generation process, based on the latent structure variable $\mathbf{z}_t$, the target word $y_t$ at the time step $t$ is drawn from a conditional probability distribution $p_\theta(\mathbf{y}_t|\mathbf{z}_t)$.
The target is to maximize the probability of each generated summary $y = \{\mathbf{y}_1, \mathbf{y}_2, \ldots, \mathbf{y}_T \}$ based on the generation process according to:
\begin{equation}
\label{integral}
{p_\theta }(y) = \prod\limits_{t = 1}^T {\int {{p_\theta }({{\mathbf{y}}_t}|{{\mathbf{z}}_t}){p_\theta }({{\mathbf{z}}_t})d{{\mathbf{z}}_t}} } 
\end{equation}
For the purpose of solving the intractable integral of the marginal likelihood as shown in Equation~\ref{integral}, a recognition model $q_\phi(\mathbf{z}_t|\mathbf{y}_{<t},\mathbf{z}_{<t})$ is introduced as an approximation to the intractable true posterior $p_\theta(\mathbf{z}_t|\mathbf{y}_{<t},\mathbf{z}_{<t})$.
The recognition model parameters $\phi$ and the generative model parameters $\theta$ can be learned jointly. The aim is to reduce the Kulllback-Leibler divergence (KL) between $q_\phi(\mathbf{z}_t|\mathbf{y}_{<t},\mathbf{z}_{<t})$ and $p_\theta(\mathbf{z}_t|\mathbf{y}_{<t},\mathbf{z}_{<t})$:
\[
\begin{split}
&{D_{KL}}[q_\phi(\mathbf{z}_t|\mathbf{y}_{<t},\mathbf{z}_{<t})\|p_\theta(\mathbf{z}_t|\mathbf{y}_{<t},\mathbf{z}_{<t})] \\ 
&= \int_z {q_\phi(\mathbf{z}_t|\mathbf{y}_{<t},\mathbf{z}_{<t})\log \frac{{q_\phi(\mathbf{z}_t|\mathbf{y}_{<t},\mathbf{z}_{<t})}}{{p_\theta(\mathbf{z}_t|\mathbf{y}_{<t},\mathbf{z}_{<t})}}dz} \\
&= {\mathbb{E}_{q_\phi(\mathbf{z}_t|\mathbf{y}_{<t},\mathbf{z}_{<t})}}[\log q_\phi(\mathbf{z}_t|\cdot) - \log p_\theta(\mathbf{z}_t|\cdot)]
\end{split}
\]
where $\cdot$ denotes the conditional variables $\mathbf{y}_{<t}$ and $\mathbf{z}_{<t}$.
Bayes rule is applied to $p_{\theta}(\mathbf{z}_t|\mathbf{y}_{<t},\mathbf{z}_{<t})$,
and we can extract $\log {p_\theta }(\mathbf{z})$ from the expectation, transfer the expectation term $\mathbb{E}_{q_\phi(\mathbf{z}_t|\mathbf{y}_{<t},\mathbf{z}_{<t})}$ back to KL-divergence, and rearrange all the terms.
Consequently the following holds:  
\begin{equation}
\begin{split}
&\log {p_\theta }(\mathbf{y}_{<t}) = \\ &{D_{KL}}[q_\phi(\mathbf{z}_t|\mathbf{y}_{<t},\mathbf{z}_{<t})\|p_\theta(\mathbf{z}_t|\mathbf{y}_{<t},\mathbf{z}_{<t})] \\
&+ {\mathbb{E}_{q_\phi(\mathbf{z}_t|\mathbf{y}_{<t},\mathbf{z}_{<t})}}[\log {p_\theta }(\mathbf{y}_{<t}|\mathbf{z}_{t})] \\
&- {D_{KL}}[q_\phi(\mathbf{z}_t|\mathbf{y}_{<t},\mathbf{z}_{<t})\|{p_\theta }(\mathbf{z}_t)]
\end{split}
\label{eq:mlhd}
\end{equation}
Let $\mathcal{L}(\theta ,\phi; y)$ represent the last two terms from the right part of Equation~\ref{eq:mlhd}:
\begin{equation}
\begin{aligned}
\mathcal{L}&(\theta ,\varphi ;y) = \\[0pt] &{\mathbb{E}_{q_\phi(\mathbf{z}_t|\mathbf{y}_{<t},\mathbf{z}_{<t})}}\big\{\sum\nolimits_{t = 1}^T\log {p_\theta }(\mathbf{y}_{t}|\mathbf{z}_{t}) \\[0pt]
&- {D_{KL}}[q_\phi(\mathbf{z}_t|\mathbf{y}_{<t},\mathbf{z}_{<t})\|{p_\theta }(\mathbf{z}_t)]\big\}
\end{aligned}
\label{eq:vlbd}
\end{equation}
Since the first KL-divergence term of Equation~\ref{eq:mlhd} is non-negative, we have $\log {p_\theta }(\mathbf{y}_{<t}) \ge \mathcal{L}(\theta ,\phi ;y)$ meaning that $\mathcal{L}(\theta ,\phi ;y)$ is a lower bound (the objective to be maximized) on the marginal likelihood. In order to differentiate and optimize the lower bound $\mathcal{L}(\theta ,\phi ;y)$, following the core idea of VAEs, we use a neural network framework for the probabilistic encoder $q_{\phi}(\mathbf{z}_t|\mathbf{y}_{<t},\mathbf{z}_{<t})$ for better approximation.

\subsection{Abstractive Summary Generation}
\label{sec:summ_gen}

We also design a neural network based framework to conduct the variational inference and generation for the recurrent generative decoder component similar to some design in previous works \cite{kingma2013auto,rezende2014stochastic,gregor2015draw}.
The encoder component and the decoder component are integrated into a unified abstractive summarization framework.
Considering that GRU has comparable performance but with less parameters and more efficient computation, we employ GRU as the basic recurrent model which updates the variables according to the following operations:
\[
\begin{array}{l}
\mathbf{r}_t = \sigma (\mathbf{W}_{xr}\mathbf{x}_t + \mathbf{W}_{hr}\mathbf{h}_{t - 1} + \mathbf{b}_r)\\
\mathbf{z}_t = \sigma (\mathbf{W}_{xz}\mathbf{x}_t + \mathbf{W}_{hz}\mathbf{h}_{t - 1} + \mathbf{b}_z)\\
\mathbf{g}_t = \tanh (\mathbf{W}_{xh}\mathbf{x}_t + \mathbf{W}_{hh}(\mathbf{r}_t \odot \mathbf{h}_{t - 1}) + \mathbf{b}_h)\\
\mathbf{h}_t = \mathbf{z}_t \odot \mathbf{h}_{t - 1} + (1 - \mathbf{z}_t) \odot \mathbf{g}_t
\end{array}
\]
where $\mathbf{r}_t$ is the reset gate, $\mathbf{z}_t$ is the update gate.
$\odot$ denotes the element-wise multiplication. $tanh$ is the  hyperbolic tangent activation function.

As shown in the left block of Figure~\ref{fig:framework}, the encoder is designed based on bidirectional recurrent neural networks.
Let $\mathbf{x}_t$ be the word embedding vector of the $t$-th word in the source sequence.
GRU maps $\mathbf{x}_t$ and the previous hidden state $\mathbf{h}_{t-1}$ to the current hidden state $\mathbf{h}_t$ in feed-forward direction and back-forward direction respectively:
\begin{equation}
\begin{array}{l}
{{\mathord{\buildrel{\lower3pt\hbox{$\scriptscriptstyle\rightharpoonup$}} 
			\over {\mathbf{h}}} }_t} = GRU({x_t},{{\mathord{\buildrel{\lower3pt\hbox{$\scriptscriptstyle\rightharpoonup$}} 
			\over {\mathbf{h}}} }_{t - 1}})\\
{{\mathord{\buildrel{\lower3pt\hbox{$\scriptscriptstyle\leftharpoonup$}} 
			\over {\mathbf{h}}} }_t} = GRU({x_t},{{\mathord{\buildrel{\lower3pt\hbox{$\scriptscriptstyle\leftharpoonup$}} 
			\over {\mathbf{h}}} }_{t - 1}})
\end{array}
\end{equation}
Then the final hidden state $\mathbf{h}_t^e \in \mathbb{R}^{2k_h}$ is concatenated using the hidden states from the two directions:
$
{\mathbf{h}}_t^e = {{\mathord{\buildrel{\lower3pt\hbox{$\scriptscriptstyle\rightharpoonup$}} 
			\over {\mathbf{h}}} }_t}||\mathord{\buildrel{\lower3pt\hbox{$\scriptscriptstyle\leftharpoonup$}} 
	\over {\mathbf{h}}} 
$. 
As shown in the middle block of Figure~\ref{fig:framework}, the decoder consists of two components: discriminative deterministic decoding  and generative latent structure modeling.

The discriminative deterministic decoding is an improved attention modeling based recurrent sequence decoder. The first hidden state $\mathbf{h}_1^d$ is initialized using the average of all the source input states: $
\mathbf{h}_1^d = \frac{1}{{{T^e}}}\sum\limits_{t = 1}^{{T^e}} {\mathbf{h}_t^e}$,
where $\mathbf{h}_t^e$ is the source input hidden state. $T^e$ is the input sequence length.
The deterministic decoder hidden state $\mathbf{h}_t^d$ is calculated using two layers of GRUs.
On the first layer, the hidden state is calculated only using the current input word embedding $\mathbf{y}_{t-1}$ and the  previous hidden state $\mathbf{h}_{t-1}^{d_1}$:
\begin{equation}
\mathbf{h}_t^{d_1} = GRU_1(\mathbf{y}_{t-1}, \mathbf{h}_{t-1}^{d_1})
\end{equation}
where the superscript $d_1$ denotes the first decoder GRU layer.
Then the attention weights at the time step $t$ are calculated based on the relationship of $\mathbf{h}_t^{d_1}$ and all the source hidden states $\{\mathbf{h}_t^e\}$. Let $a_{i,j}$ be the attention weight between $\mathbf{h}_i^{d_1}$ and $\mathbf{h}_j^{e}$, which can be calculated using the following formulation:
\[
\begin{aligned}
{a_{i,j}} &= \frac{{\exp ({e_{i,j}})}}{{\sum\nolimits_{j' = 1}^{{T^e}} {\exp ({e_{i,j'}})} }}\\
{e_{i,j}} &= {\mathbf{v}^T}\tanh (\mathbf{W}_{hh}^d\mathbf{h}_i^{{d_1}} + \mathbf{W}_{hh}^e\mathbf{h}_j^e + {\mathbf{b}_a})
\end{aligned}
\]
where $\mathbf{W}_{hh}^d \in \mathbb{R}^{k_h \times k_h}$, $\mathbf{W}_{hh}^e \in \mathbb{R}^{k_h \times 2k_h}$, $\mathbf{b}_a \in \mathbb{R}^{k_h}$, and $\mathbf{v} \in \mathbb{R}^{k_h}$.
The attention context is obtained by the weighted linear combination of all the source hidden states:
\begin{equation}
{\mathbf{c}_t} = \sum\nolimits_{j' = 1}^{{T^e}} {{a_{t,j'}}\mathbf{h}_{j'}^e} 
\end{equation}

The final deterministic hidden state $\mathbf{h}_t^{d_2}$ is the output of the second decoder GRU layer, jointly considering the word $\mathbf{y}_{t-1}$, the previous hidden state $\mathbf{h}_{t-1}^{d_2}$, and the attention context $\mathbf{c}_t$:
\begin{equation}
\mathbf{h}_t^{d_2} = GRU_2(\mathbf{y}_{t-1}, \mathbf{h}_{t-1}^{d_2}, \mathbf{c}_t)
\end{equation}

For the component of recurrent generative model, inspired by some ideas in previous works \cite{kingma2013auto,rezende2014stochastic,gregor2015draw}, we assume that both the prior and posterior of the latent variables are Gaussian, i.e.,  $p_\theta (\mathbf{z}_t) = \mathcal{N}(0, \mathbf{I})$ and $q_{\phi}(\mathbf{z}_t|\mathbf{y}_{<t}, \mathbf{z}_{<t}) = \mathcal{N}(\mathbf{z}_t; \boldsymbol{\mu}, \boldsymbol{\sigma}^2\mathbf{I})$, where  $\boldsymbol{\mu}$ and $\boldsymbol{\sigma}$ denote the variational mean and standard deviation respectively, which can be calculated via a multilayer perceptron. 
Precisely, given the word embedding $\mathbf{y}_{t-1}$, the previous latent structure variable $\mathbf{z}_{t-1}$, and the previous deterministic hidden state $\mathbf{h}_{t-1}^{d}$, we first project it to a new hidden space:
\[
{\mathbf{h}_{t}^{e_z}} = g({\mathbf{W}_{yh}^{e_z}}\mathbf{y}_{t-1} +
{\mathbf{W}_{zh}^{e_z}}\mathbf{z}_{t-1} + {\mathbf{W}_{hh}^{e_z}}\mathbf{h}^d_{t-1} + {\mathbf{b}^{e_z}_h})
\]
where $\mathbf{W}_{yh}^{e_z} \in \mathbb{R}^{k_h \times k_w}$, $\mathbf{W}_{zh}^{e_z} \in \mathbb{R}^{k_h \times k_z}$, $\mathbf{W}_{hh}^{e_z} \in \mathbb{R}^{k_h \times k_h}$, and $\mathbf{b}^{e_z}_h \in \mathbb{R}^{k_h}$. $g$ is the sigmoid activation function: $\sigma(\mathbf{x}) = 1 / (1 + e^{-\mathbf{x}})$.
Then the Gaussian parameters $\boldsymbol{\mu}_t \in \mathbb{R}^{k_z}$ and $\boldsymbol{\sigma}_t \in \mathbb{R}^{k_z}$ can be obtained via a linear transformation based on  $\mathbf{h}_t^{e_z}$:
\begin{equation}
\begin{aligned}
\boldsymbol{\mu}_t  &= {\mathbf{W}_{h\mu }^{e_z}}{\mathbf{h}_t^{e_z}} + {\mathbf{b}_{\mu }^{e_z}}\\
\log ({\boldsymbol{\sigma}_t ^2}) &= {\mathbf{W}_{h\sigma }}{\mathbf{h}_t^{e_z}} + {\mathbf{b}_{\sigma }^{e_z}}\\
\end{aligned}
\end{equation}
The latent structure variable $\mathbf{z}_t \in \mathbb{R}^{k_z}$ can be calculated using the reparameterization trick:
\begin{equation}
\label{e:ptrick}
\begin{array}{l}
\boldsymbol{\varepsilon}  \sim \mathcal{N}(\mathbf{0}, \mathbf{I}), \ \
\mathbf{z}_t = \boldsymbol{\mu}_t  + \boldsymbol{\sigma}_t  \otimes \varepsilon \\
\end{array}
\end{equation}
where $\boldsymbol{\varepsilon} \in \mathbb{R}^{k_z}$ is an auxiliary noise variable. The process of inference for finding $\mathbf{z}_t$ based on neural networks can be teated as a variational encoding process.

To generate summaries precisely, we first integrate the recurrent generative decoding component with the discriminative deterministic decoding component, and map the latent structure variable $\mathbf{z}_t$ and the deterministic decoding hidden state $\mathbf{h}_t^{d_2}$ to a new hidden variable:
\begin{equation}
{\mathbf{h}_{t}^{d_y}} = tanh({\mathbf{W}_{zh}^{d_y}}\mathbf{z}_{t} +
{\mathbf{W}_{hh}^{d_z}}\mathbf{h}_{t}^{d_2} + {\mathbf{b}^{d_y}_h})
\end{equation}

Given the combined decoding state $\mathbf{h}_{t}^{d_y}$ at the time $t$, the probability of generating any target word $y_t$ is given as follows:
\begin{equation}
\mathbf{y}_t = \varsigma({\mathbf{W}_{hy}^{d}}\mathbf{h}_{t}^{d_y} + {\mathbf{b}^{d}_{hy}})
\end{equation}
where ${\mathbf{W}_{hy}^{d}} \in \mathbb{R}^{k_y \times k_h}$ and ${\mathbf{b}^{d}_{hy}} \in \mathbb{R}^{k_y}$. $\varsigma(\cdot)$ is the softmax function.
Finally, we use a beam search algorithm \cite{koehn2004pharaoh} for decoding and generating the best summary.

\subsection{Learning}
Although the proposed model contains a recurrent generative decoder, the whole framework is fully differentiable. As shown in Section~\ref{sec:summ_gen}, both the recurrent deterministic decoder and the recurrent generative decoder are designed based on neural networks. Therefore, all the parameters in our model can be optimized in an end-to-end paradigm using back-propagation.
We use  $\{X\}_N$ and $\{Y\}_N$ to denote the training source and target sequence.
Generally, the objective of our framework consists of two terms.
One term is the negative log-likelihood of the generated summaries, and the other one is the variational lower bound $\mathcal{L}(\theta ,\phi ;Y)$ mentioned in Equation~\ref{eq:vlbd}.
Since the variational lower bound $\mathcal{L}(\theta ,\phi ;Y)$ also contains a likelihood term, we can merge it with the likelihood term of summaries.
The final objective function, which needs to be minimized, is formulated as follows:
\begin{equation}
\small
\begin{split}
\mathcal{J} = \frac{1}{N}\sum\limits_{n = 1}^N \sum\limits_{t = 1}^T  \Bigg\{ -\log \bigg[p(y_t^{(n)}|y_{ < t}^{(n)},{X^{(n)}})\bigg] \\
+ {D_{KL}}\bigg[{q_\phi }(\mathbf{z}_t^{(n)}|\mathbf{y}_{ < t}^{(n)},\mathbf{z}_{ < t}^{(n)})\|{p_\theta }(\mathbf{z}_t^{(n)})\bigg] \Bigg\} 
\end{split}
\label{eq:obj}
\end{equation}

\section{Experimental Setup}

\subsection{Datesets}
We train and evaluate our framework on three popular datasets.
\textbf{Gigawords} is an English sentence summarization dataset prepared based on Annotated Gigawords\footnote{https://catalog.ldc.upenn.edu/ldc2012t21} by extracting the first sentence from articles with the headline to form a source-summary pair.
We directly download the prepared dataset  used in \cite{rush2015neural}.
It roughly contains 3.8M training pairs, 190K validation pairs, and 2,000 test pairs.
\textbf{DUC-2004}\footnote{http://duc.nist.gov/duc2004} is another English dataset only used for testing in our experiments. It contains 500 documents. Each document contains 4 model summaries written by experts. The length of the summary is limited to 75 bytes.
\textbf{LCSTS} is a large-scale Chinese short text summarization dataset, consisting of pairs of (short text, summary) collected from Sina Weibo\footnote{http://www.weibo.com} \cite{hu2015lcsts}.
We take Part-I as the training set, Part-II as the development set, and Part-III as the test set. There is a score in range $1\sim5$ labeled by human to indicate how relevant an article and its summary is. We only reserve those pairs with scores no less than 3. The size of the three sets are 2.4M, 8.7k, and 725 respectively.
In our experiments, we only take Chinese character sequence as input, without performing word segmentation.

\subsection{Evaluation Metrics}
We use ROUGE score \cite{lin2004rouge} as our evaluation metric with standard options.
The basic idea of ROUGE is to count the number of overlapping units between generated summaries and the reference summaries, such as overlapped n-grams, word sequences, and word pairs.
F-measures of ROUGE-1 (R-1), ROUGE-2 (R-2), ROUGE-L (R-L) and ROUGE-SU4 (R-SU4) are reported.


\subsection{Comparative Methods}
We compare our model with some baselines and state-of-the-art methods.
Because the datasets are quite standard, so we just extract the results from their papers. Therefore the baseline methods on different datasets may be slightly different.

\begin{itemize}
	
	\item \textbf{TOPIARY} \cite{zajic2004bbn} is the best on DUC2004 Task-1 for compressive text summarization.
	It combines a system using linguistic based transformations and an unsupervised topic detection algorithm for compressive text summarization.
	
	\item \textbf{MOSES+} \cite{rush2015neural} uses a phrase-based statistical machine translation system trained on Gigaword to produce summaries.
	It also augments the phrase table with ``deletion'' rulesto improve the baseline performance, and MERT is also used to improve the quality of generated summaries.
	
	\item \textbf{ABS} and \textbf{ABS+} \cite{rush2015neural} are both the neural network based models with local attention modeling for abstractive sentence summarization.
	ABS+ is trained on the Gigaword corpus, but combined with an additional log-linear extractive summarization model with handcrafted features.
	
	\item \textbf{RNN} and \textbf{RNN-context} \cite{hu2015lcsts} are two seq2seq architectures. RNN-context integrates attention mechanism to model the context.
	
	\item \textbf{CopyNet} \cite{gu2016incorporating} integrates a copying mechanism into the sequence-to-sequence framework.
	
	\item \textbf{RNN-distract} \cite{chen2016distraction} uses a new attention mechanism by distracting the historical attention in the decoding steps.
	
	\item \textbf{RAS-LSTM} and \textbf{RAS-Elman} \cite{chopra2016abstractive} both consider words and word positions as input and use convolutional encoders to handle the source information.
	For the attention based sequence decoding process, RAS-Elman selects Elman RNN \cite{elman1990finding} as decoder, and RAS-LSTM selects Long Short-Term Memory architecture \cite{hochreiter1997long}.
	
	\item \textbf{LenEmb} \cite{kikuchi2016controlling} uses a mechanism to control the summary length by considering the length embedding vector as the input.
	
	\item \textbf{ASC+FSC$_1$} \cite{miao2016language} uses a generative model with attention mechanism to conduct the sentence compression problem.
	The model first draws a latent summary sentence from a background language model, and then subsequently draws the observed sentence conditioned on this latent summary.
	
	\item \textbf{lvt2k-1sent} and \textbf{lvt5k-1sent} \cite{nallapati2016abstractive} utilize a trick to control the vocabulary size to improve the training efficiency.

\end{itemize}

\subsection{Experimental Settings}
For the experiments on the English dataset Gigawords, we set the dimension of word embeddings to 300, and the dimension of hidden states and latent variables to 500.
The maximum length of documents and summaries is 100 and 50 respectively.
The batch size of mini-batch training is 256.
For DUC-2004, the maximum length of summaries is 75 bytes.
For the dataset of LCSTS, the dimension of word embeddings is 350.
We also set the dimension of hidden states and latent variables to 500.
The maximum length of documents and summaries is 120 and 25 respectively, and the batch size is also 256. 
The beam size of the decoder was set to be 10.
Adadelta \cite{schmidhuber2015deep} with hyperparameter $\rho = 0.95$ and $\epsilon = 1e-6$ is used for gradient based optimization.
Our neural network based framework is implemented using Theano \cite{2016arXiv160502688short}.


\section{Results and Discussions}

\subsection{ROUGE Evaluation}

\begin{table}[!htb]
	\centering
	\caption{ROUGE-F1 on validation sets}
	\label{tab:rouge-stand}
	\begin{tabular}{l c c c c}
		\hline
		\textbf{Dataset} & \textbf{System}  & \textbf{R-1} & \textbf{R-2} & \textbf{R-L} \\
		\hline
		GIGA  & StanD       & 32.69 & 15.29 & 30.60  \\
		& DRGD & \textbf{36.25} & \textbf{17.61} & \textbf{33.55}  \\
		\hline
		LCSTS & StanD       & 33.88 & 21.49 & 31.05  \\
		& DRGD      & \textbf{36.71} & \textbf{24.00} & \textbf{34.10}  \\
		\hline
	\end{tabular}
\end{table}

We first depict the performance of our model DRGD by comparing to the standard decoders (StanD) of our own implementation.
The comparison results on the validation datasets of Gigawords and LCSTS are shown in Table~\ref{tab:rouge-stand}.
From the results we can see that our proposed generative decoders DRGD can obtain obvious improvements on abstractive summarization than the standard decoders.
Actually, the performance of the standard decoders is similar with those mentioned popular baseline methods.

\begin{table}[!t]
	\centering
	\caption{ROUGE-F1 on Gigawords}
	\label{tab:rouge-agiga}
	\begin{tabular}{p{2.6cm} c c c}
		\hline
		\textbf{System}  & \textbf{R-1} & \textbf{R-2} & \textbf{R-L}  \\
		\hline
		ABS       & 29.55 & 11.32 & 26.42  \\
		ABS+       & 29.78 & 11.89 & 26.97  \\
		RAS-LSTM       & 32.55 & 14.70 & 30.03  \\
		RAS-Elman       & 33.78 & 15.97 & 31.15  \\
		ASC + FSC$_1$       & 34.17 & 15.94 & 31.92  \\
		lvt2k-1sent     & 32.67 & 15.59 & 30.64  \\
		lvt5k-1sent     & 35.30 & 16.64 & 32.62  \\
		\textbf{DRGD}       & \textbf{36.27} & \textbf{17.57} & \textbf{33.62}  \\
		\hline
	\end{tabular}
\end{table}

\begin{table}[!t]
	\centering
	\caption{ROUGE-Recall on DUC2004}
	\label{tab:rouge-duc04}
	\begin{tabular}{p{2.6cm} c c c}
		\hline
		\textbf{System}  & \textbf{R-1} & \textbf{R-2} & \textbf{R-L} \\
		\hline
		TOPIARY & 25.12 & 6.46 & 20.12  \\
		MOSES+ & 26.50 & 8.13 & 22.85  \\
		ABS       & 26.55 &	7.06 &	22.05  \\
		ABS+       & 28.18 & 8.49 & 23.81  \\
		RAS-Elman       & 28.97 & 8.26 & 24.06  \\
		RAS-LSTM       & 27.41 & 7.69 & 23.06  \\
		LenEmb       & 26.73 & 8.39 & 23.88  \\
		lvt2k-1sen       & 28.35 & 9.46 & 24.59  \\
		lvt5k-1sen & 28.61 & 9.42 & 25.24  \\
		\textbf{DRGD}       & \textbf{31.79} & \textbf{10.75} & \textbf{27.48} \\
		\hline
	\end{tabular}
\end{table}

\begin{table}[!t]
	\centering
	\caption{ROUGE-F1 on LCSTS}
	\label{tab:rouge-lcsts}
	\begin{tabular}{p{2.6cm} c c c}
		\hline
		\textbf{System}  & \textbf{R-1} & \textbf{R-2} & \textbf{R-L} \\
		\hline
		RNN       & 21.50 & 8.90 & 18.60  \\
		RNN-context       & 29.90 & 17.40 & 27.20  \\
		CopyNet       & 34.40 & 21.60 & 31.30  \\
		RNN-distract & 35.20 & 22.60 & 32.50  \\
		\textbf{DRGD}       & \textbf{36.99} & \textbf{24.15} & \textbf{34.21} \\
		\hline
	\end{tabular}
\end{table}

The results on the English datasets of Gigawords and DUC-2004 are shown in Table~\ref{tab:rouge-agiga} and Table~\ref{tab:rouge-duc04} respectively.
Our model DRGD achieves the best summarization performance on all the ROUGE metrics.
Although ASC+FSC$_1$ also uses a generative method to model the latent summary variables, the representation ability is limited and it cannot bring in noticeable improvements. 
It is worth noting that the methods lvt2k-1sent and lvt5k-1sent \cite{nallapati2016abstractive} utilize linguistic features such as parts-of-speech tags, named-entity tags, and TF and IDF statistics of the words as part of the document representation.
In fact, extracting all such features is a time consuming work, especially on  large-scale datasets such as Gigawords.  lvt2k and lvt5k are not end-to-end style models and are more complicated than our model in practical applications.

The results on the Chinese dataset LCSTS are shown in Table~\ref{tab:rouge-lcsts}.
Our model DRGD also achieves the best performance. Although CopyNet employs a copying mechanism to improve the summary quality and RNN-distract considers attention information diversity in their decoders, our model is still better than those two methods demonstrating that the latent structure information learned from target summaries indeed plays a role in abstractive summarization.
We also believe that integrating the copying mechanism and coverage diversity in our framework will further improve the summarization performance. 


\subsection{Summary Case Analysis}
In order to analyze the reasons of improving the performance, we compare the generated summaries by DRGD and the standard decoders StanD used in some other works such as \cite{chopra2016abstractive}.
The source texts, golden summaries, and the generated summaries are shown in Table~\ref{tab:latent_structure}.
From the cases we can observe that DRGD can indeed capture some latent structures which are consistent with the golden summaries. For example, our result for S(1) ``\textit{Wuhan wins men's soccer title at Chinese city games}'' matches the ``Who Action What'' structure.
However, the standard decoder StanD ignores the latent structures and generates some loose sentences, such as the results for S(1) ``\textit{Results of men's volleyball at Chinese city games}'' does not catch the main points.
The reason is that the recurrent variational auto-encoders used in our framework  have better representation ability and can capture more effective and complicated latent structures from the sequence data.
Therefore, the summaries generated by DRGD have consistent latent structures with the ground truth, leading to a better ROUGE evaluation.

\begin{table}[!t]
	\centering
	
	\caption{Examples of the generated summaries.}
	\label{tab:latent_structure}
	\begin{tabular}{p{7.2cm}}
		\hline
		\hline
		\textbf{S(1)}: hosts wuhan won the men 's soccer title by beating beijing shunyi \#-\# here at the \#th chinese city games on friday.\\
		\textbf{Golden}: hosts wuhan wins men 's soccer title at chinese city games.\\
		\textbf{StanD}: results of men 's volleyball at chinese city games.\\
		\textbf{DRGD}: \textbf{wuhan wins men 's soccer title at chinese city games.}\\
		\hline
		\textbf{S(2)}: UNK and the china meteorological administration tuesday signed an agreement here on long - and short-term cooperation in projects involving meteorological satellites and satellite meteorology.\\
		\textbf{Golden}: UNK china to cooperate in meteorology.\\
		\textbf{StanD}: weather forecast for major chinese cities.\\
		\textbf{DRGD}: \textbf{china to cooperate in meteorological satellites.}\\
		\hline
		\textbf{S(3)}: the rand gained ground against the dollar at the opening here wednesday , to \#.\# to the greenback from \#.\# at the close tuesday.\\
		\textbf{Golden}: rand gains ground.\\
		\textbf{StanD}: rand slightly higher against dollar.\\
		\textbf{DRGD}: \textbf{rand gains ground against dollar.}\\
		\hline
		\textbf{S(4)}: new zealand women are having more children and the country 's birth rate reached its highest level in \#\# years , statistics new zealand said on wednesday.\\
		\textbf{Golden}: new zealand birth rate reaches \#\#-year high.\\
		\textbf{StanD}: new zealand women are having more children birth rate hits highest level in \#\# years.\\
		\textbf{DRGD}: \textbf{new zealand 's birth rate hits \#\#-year high.}\\
		\hline
		\hline
	\end{tabular}
\end{table}

\section{Conclusions}
We propose a deep recurrent generative decoder (DRGD) to improve the abstractive summarization performance.
The model is a sequence-to-sequence oriented encoder-decoder framework equipped with a latent structure modeling component. 
Abstractive summaries are generated based on both the latent variables and the deterministic states. 
Extensive experiments on benchmark datasets show that DRGD achieves improvements over the state-of-the-art methods.

\bibliography{emnlp2017}
\bibliographystyle{emnlp_natbib}

\end{document}